\documentclass[11pt]{article}
\usepackage[final]{acl}   

\usepackage{amsmath,amsfonts,bm}

\def\eqref#1{equation~\ref{#1}}
\def\1{\bm{1}}

\DeclareMathAlphabet{\mathsfit}{\encodingdefault}{\sfdefault}{m}{sl}
\SetMathAlphabet{\mathsfit}{bold}{\encodingdefault}{\sfdefault}{bx}{n}

\usepackage{hyperref}
\usepackage{url}
\usepackage{tabularx}
\usepackage{booktabs}
\usepackage{wrapfig}
\usepackage{graphicx}
\usepackage{multirow}

\usepackage{array}
\usepackage{caption}
\usepackage{siunitx}

\title{Motivating Next-Gen Accelerators with Flexible \(N:M\) Activation Sparsity via Benchmarking Lightweight Post-Training Sparsification Approaches}  

\author{
 Shirin Alanova$^\clubsuit$ \textsuperscript{7}, 
 Kristina Kazistova$^\clubsuit$ \textsuperscript{7},
 Ekaterina Galaeva$^\clubsuit$ \textsuperscript{7}, 
 Alina Kostromina$^\clubsuit$ \textsuperscript{4,3}, \\
 Vladimir Smirnov \textsuperscript{6},
 Dmitry Redko \textsuperscript{1}, 
 Alexey Dontsov \textsuperscript{1}, \\
 Maxim Zhelnin \textsuperscript{5},
 Evgeny Burnaev \textsuperscript{1,2}, 
 Egor Shvetsov \textsuperscript{1}
\\
 \textsuperscript{1} \small{Applied AI} 
 \textsuperscript{2} \small{Artificial Intelligence Research Institute} 
 \textsuperscript{3} \small{HSE University}
 \textsuperscript{4} \small{Sb AI Lab} \\
 \textsuperscript{5} \small{MWS AI}
 \textsuperscript{6} \small{Yandex}
 \textsuperscript{7} \small{Independent Researcher}
\\
  \small{ $^\clubsuit$ indicates equal contribution.} \\
}

\begin{document}

\maketitle
% \begin{abstract}
% \noindent The demand for efficient large language model inference has spurred interest in sparsification, yet current hardware support remains narrowly focused on 2:4 weight sparsity. In this work, we argue that \textit{activation sparsity} despite being overlooked in hardware design offers a more promising path for dynamic, input-adaptive compression with significant I/O and memory benefits. We present a comprehensive post-training study of $N{:}M$ activation pruning in four LLMs (Llama2-7B, Llama3.1-8.1B, Qwen2.5-7B, Gemma3-4B), demonstrating that activation pruning consistently outperforms weight pruning at matched sparsity levels. We evaluate lightweight plug-and-play error mitigation and selection strategies that require minimal or no calibration data with four sparsity patterns 2:4, 4:8, 8:16, and 16:32, with 16:32 approaching unstructured 50\% sparsity performance and being 3$\times$ better than 2:4, while 8:16 offers an optimal balance of accuracy and practicality. Our results provide strong algorithmic evidence that next-generation accelerators should consider native support of $N{:}M$ activation sparsity. The code is available \href{https://anonymous.4open.science/r/Structured-Sparse-Activations-Inference-EC3C/README.md}{here}.
% \end{abstract}

\begin{abstract}
\noindent The demand for efficient large language model inference has spurred interest in sparsification, yet current hardware support remains narrowly focused on 2:4 weight sparsity. In this work, we argue that activation sparsity despite being overlooked in hardware design offers a promising path for dynamic, input-adaptive compression with significant I/O and memory benefits. We present a comprehensive post-training study of $N{:}M$ activation pruning across four LLMs (Llama2-7B-chat, Llama3.1-8B-Instruct, Qwen2.5-7B-Instruct, Gemma3-4B-Instruct), demonstrating that activation pruning consistently outperforms weight pruning at matched sparsity levels. We evaluate lightweight, plug-and-play error mitigation and selection strategies that require minimal or no calibration data across four sparsity patterns: 2:4, 4:8, 8:16, and 16:32. Among these, 16:32 approaches the performance of unstructured 50\% sparsity and is is approximately 2.7$\times$ better than 2:4, while 8:16 offers an optimal balance of accuracy and practicality. Our results provide evidence that next-generation accelerators should consider native support for $N{:}M$ activation sparsity and can serve as a strong baseline for the future methods. The code is available \href{https://anonymous.4open.science/r/Structured-Sparse-Activations-Inference-EC3C/README.md}{here}.
\end{abstract}

\section{Introduction}
\label{sec:introduction}

% The expanding capabilities of Large Language Models (LLMs) have driven enormous demand for efficient AI inference. Accelerating inference typically involves either reducing numerical precision via quantization or sparsification to reduce the number of parameters \cite{0.1.1}. While sparsity improves efficiency by reducing computation and I/O traffic, current hardware implementations present a critical bottleneck. Only NVIDIA GPUs currently support structured N:M sparsity, specifically the fixed 2:4 pattern, which achieves modest 1.5–1.7x inference acceleration for 7B models \cite{0.1.3}.

% This limitation means that the full potential of N:M sparsity for activations remains largely untapped. While dynamic activation sparsity preserves model capacity better than static weight pruning, the overhead of applying non-standard patterns on current general-purpose hardware is prohibitive. Therefore, this work aims to shift the focus from immediate practical deployment of arbitrary N:M patterns to providing the necessary empirical evidence to motivate future hardware design. We focus on two key areas: first, evaluating lightweight, post-training error mitigation strategies that can work within current constraints, and second, demonstrating the significant quality improvements offered by more flexible patterns like 8:16, which should be considered for next-generation accelerators.

Large Language Models (LLMs) have intensified demand for efficient inference. A common rule of thumb suggests serving speed for a dense $N$-parameter model scales as $\propto 1/\sqrt{N}$~\citep{erdil2025inference}. Inference is often accelerated via quantization~\citep{frantar2022gptq, 10639389} or sparsification~\citep{frantar2023sparsegpt,maximov20252}, with sparsity reducing both compute and memory I/O.

\textbf{Weights vs. Activations.} Although weight and activation sparsity yield the same theoretical FLOP count, they differ in practice. Weight sparsity enables static compression but can irreversibly degrade model quality, whereas activation sparsity is input-adaptive and better preserves model capacity though it requires an additional pruning step for each input.

\textbf{Accelerating LLMs with Sparse Activations.} Naturally, pruning is most effective when the values to be pruned already have low magnitudes or are zero, which is often the case for some LLMs' intermediate representations~\citep{liu2023deja, li2022lazy}. Activation sparsity was employed to accelerate the decoding stage by enabling faster \emph{sparse vector}–\emph{dense matrix} multiplications up to 2× speedup using specialized kernels~\citep{song2024turbo, song2024powerinfer, liu2024training, lee2024cats}. These gains are most pronounced for batch size 1 and diminish as batch size increases~\citep{shrestha2025polar}. The speedup primarily arises because rows of the dense matrix corresponding to zero elements in the sparse vector can be skipped during computation. Moreover, these methods often require some predictive mechanism to upload required weight indices into memory ahead of time~\cite{liu2023deja}.

% most methods rely on sparsity induced by MLP activations~\citep{mirzadeh2023relu}.

% \textbf{Semi-structured N:M activation sparsity} (keeping $N$ non-zeros per block of size $M$) can extend benefits beyond single vector--matrix products by balancing unstructured accuracy retention~\citep{zhu2016trained, paul2022unmasking} with structured hardware efficiency~\citep{liu2017learning, molchanov2019importance, hubara2021accelerated}. Although 2:4 activation sparsity has been explored for training~\citep{haziza2025accelerating, wang2024q}, and post-training weight sparsification~\cite{maximov20252, NIPS2015_ae0eb3ee, frantar2023sparsegpt, kurtic2023ziplm} post-training N:M activation pruning is largely unexplored since it is not widely supported on the current hardware.

\textbf{Semi-structured N:M activation sparsity}, keeping $N$ non-zeros per block of size $M$, can extend benefits beyond single vector--matrix products by employing hardware support. Although 2:4 activation sparsity has been explored for training~\citep{haziza2025accelerating, wang2024q}, and post-training weight sparsification~\cite{maximov20252, NIPS2015_ae0eb3ee, frantar2023sparsegpt, kurtic2023ziplm} post-training N:M activation pruning is largely unexplored because it is not supported on commercial hardware.

\textbf{Hardware and $N{:}M$ Sparsity.}
Current commercial hardware provides native support only for 2:4 structured sparsity in weights, which can deliver $\sim$1.5--1.7$\times$ inference speedup and $\sim$1.5$\times$ energy reduction for models like 7B LLMs, primarily by halving memory bandwidth demand~\citep{fang2024maskllm, lin2023efficient}. However, this fixed pattern offers limited flexibility: a 2:4 block has only $\binom{4}{2}=6$ valid configurations. In contrast, larger patterns like 8:16 provide the same 2$\times$ bandwidth reduction but with $\binom{16}{8}=12{,}870$ possible layouts nearly 10$\times$ more than four concatenated 2:4 blocks ($6^4=1{,}296$)---at a modest increase in metadata cost (from $\approx$0.75 to $\approx$0.875 bits per element). This highlights a significant opportunity for more flexible sparsity. Although the dynamic nature of activation sparsity requires additional computations and is impractical on existing hardware, recent developments point towards its future feasibility. We estimate potential hardware overhead to support dynamic sparsity and feasibility of development in Appendix~\ref{app:hardware}.

\textbf{Focus and Motivation.}
The primary goal of this work is twofold: to motivate the development of hardware that natively supports semi-structured activation sparsity, and to benchmark existing approaches for inducing such sparsity in activations. Generally, these approaches address two key challenges:  

\textbf{(P1) Selection Strategy.} As with weight sparsification, the choice of which activations to retain critically affects model accuracy~\citep{zhelnin2025gift}.  

\textbf{(P2) Error Mitigation.} While post-training fine-tuning can recover lost performance, it is often impractical due to computational cost or risks to safety alignment~\citep{kharinaev2025investigating}. We therefore mainly focus on lightweight, \textit{plug-and-play} methods that require minimal (e.g., WikiText) or zero calibration data.

\paragraph{Our Contributions:}
\begin{itemize}
    \item First, we demonstrate that \textbf{activation sparsity outperforms weight sparsity}. At matched sparsity levels, activation pruning consistently yields higher accuracy than weight pruning across four diverse large language models Llama2-7B-chat, Llama3.1-8B-Instruct, Qwen2.5-7B-Instruct and Gemma3-4B-Instruct highlighting activations as a more promising target for future sparse accelerators.
    \item Second, we benchmark four plug-and-play error mitigation techniques, three of which are applied to semi-structured activation sparsity for the first time, including statistical approaches such as median shift and variance correction. We also evaluate three selection criteria.  These methods establish strong, retraining-free baselines that require minimal metadata, aligning well with hardware constraints.

\item Finally, we analyze a wide range of structured sparsity patterns and show that larger patterns dramatically improve model fidelity: the 16:32 pattern achieves performance close to unstructured 50\% sparsity and retains over three times more accuracy than the conventional 2:4 pattern. Nevertheless, considering implementation trade-offs, we advocate for \textbf{8:16} as the optimal balance it offers twice the accuracy retention of 2:4 while remaining highly practical for near-term hardware adoption.

\end{itemize}

Together, these results provide concrete evidence that expanding hardware support beyond static 2:4 weight sparsity can unlock significant efficiency gains without compromising model quality, thereby motivating the next generation of sparsity-aware architectures.

\vspace{-0.5em}

\begin{table*}[ht]
\centering
\caption{Brief description of evaluated activation \emph{pruning metrics} (top) and \emph{transformations} (bottom). Abbreviations in \textbf{bold} with an asterisk (*) denote methods proposed here or first evaluated with sparse activations.}
\resizebox{\textwidth}{!}{%
\begin{tabular}{@{}l p{4cm} p{13cm}@{}}
\toprule
\textbf{Short Name} & \textbf{Method} & \textbf{Key Mechanism} \\
\midrule
\multicolumn{3}{@{}l}{\textit{Pruning metrics}}\\
ACT & Magnitude Pruning & Selects based on activation magnitude \\
WT & Weight-based Pruning & Selects by corresponding weight magnitude \\
\textbf{CLACT*} & Cosine Loss Activation & A metric inspired by cosine similarity from~\cite{mi2025ace} \\
Amber-Pruner~(\citeyear{an2025amberprunerleveragingnm}) & Weights Importance & A metric which  accounts for important weights after outlier removal and normalization \\

\midrule
\multicolumn{3}{@{}l}{\textit{Transformations}}\\
%PCS & Per-Channel Smoothing & Smoothing before pruning inspired by SmoothQuant~\citep{xiao2023smoothquant} \\
\textbf{D-PTS*} & Dynamic Per-Token Shift & Batch-wise dynamic centering of activations before sparsification \\
S-PTS~(\citeyear{chua2024shiftbias}) & Static Per-Token Shift & Fixed centering of activations before sparsification using a per-token bias value \textbf{pre-collected on WikiText-2} \\
\textbf{L-PTS*} & Learnable Per-Token Shift & Fixed centering of activations before sparsification using per-token bias value \textbf{learned on WikiText-2} \\
\textbf{VAR*} & Variance Correction & Token-wise variance normalization after sparsification  \\
\textbf{VAR+L-PTS*} & Scaling + Learnable Shift & Apply VAR scaling, then add per-token bias value \textbf{learned on WikiText-2} \\
R-Sparse~(\citeyear{kamirul2025rsparsercnnsarship}) & Rank-Aware Sparsity & Combines sparse activations with weight low-rank SVD factors \textbf{learned on WikiText-2} \\
\bottomrule
\end{tabular}%
}
\label{tab:methods_table}
\end{table*}

\section{Methods}
In the sections below, we formally define the pruning criteria and error mitigation strategies evaluated in this work. A brief summary of these methods is provided in Table~\ref{tab:methods_table}.
% \begin{table}[t]
% \centering
% \caption{Summary of evaluated pruning metrics and transformations. Full descriptions are provided in Appendix~\ref{tab:methods_table}.}
% \label{tab:strategies_descr}
% \footnotesize
% \setlength{\tabcolsep}{4.2pt}
% \renewcommand{\arraystretch}{0.92}

% \begin{tabular}{@{}ll@{}}
% \toprule
% \textbf{Short name} & \textbf{Description} \\
% \midrule

% \multicolumn{2}{@{}l}{\textbf{Pruning metrics}}\\
% ACT & Activation magnitude \\
% WT & Weight magnitude \\
% CLACT* & Context-aware cosine-loss-inspired \\
% Amber-Pruner & Weight-informed importance \\

% \midrule
% \multicolumn{2}{@{}l}{\textbf{Transformations}}\\
% D-PTS* & Dynamic per-token shift \\
% S-PTS & Static per-token shift \\
% L-PTS* & Learnable per-token shift \\
% VAR* & Variance correction \\
% VAR+L-PTS* & VAR + learnable shift \\
% R-Sparse & Sparse activations + low-rank weights \\

% \bottomrule
% \end{tabular}
% \end{table}

\subsection{Preliminaries}
For a linear layer with weights $\mathbf{W}$, input activations $\mathbf{X}$, and output $\mathbf{Y}$:
\begin{equation}
\mathbf{Y}=\mathbf{X}\mathbf{W}^\top.
\end{equation}
We construct a sparsity mask $\mathbf{M}$ using a metric $S$ and threshold $t$:
\begin{equation}
\mathbf{M}_{ij}=
\begin{cases}
1, & S(\mathbf{X}_{ij})\ge t,\\
0, & S(\mathbf{X}_{ij})< t,
\end{cases}
\end{equation}
yielding
\begin{equation}
\mathbf{Y}_{p}=(\mathbf{X}\odot\mathbf{M})\mathbf{W}^\top.
\end{equation}

Unstructured sparsity applies a global threshold over all elements. Semi-structured $N{:}M$ sparsity partitions rows (or columns) into non-overlapping blocks of size $M$ and keeps the top-$N$ elements per block by $S$ (e.g., 2{:}4 removes 50\%)~\citep{hu2024accelerating}. We also study 8{:}16, which retains the same 50\% density with higher flexibility and modest metadata overhead~\citep{maximov20252}.

\subsection{Pruning Criterion}
\label{sec:pruning_criteria}

\textbf{ACT:} This is the magnitude activation pruning metric (\textbf{ACT}), defined as the absolute value of the element $\mathbf{X}_{ij}: S_{\textit{ACT}}(\mathbf{X}_{ij}) = |\mathbf{X}_{ij}|$.

\textbf{WT:} This is the weight-based pruning metric (\textbf{WT}), defined as the absolute value of the corresponding weight $\mathbf{W}_{ij}$: $S_{\textit{WT}}(\mathbf{W}_{ij}) = |\mathbf{W}_{ij}|$. 

\textbf{CLACT:} Inspired by output cosine similarity in~\citep{mi2025ace}, we propose Cosine Loss ACTivation (CLACT), a context-aware score that emphasizes activations aligned with their row/column energy:
\begin{equation}
S_{\textit{CLACT}}(\mathbf{X}_{ij})=
\dfrac{|\mathbf{X}_{ij}|}{\sqrt{\sum_{k=1}^{h}\mathbf{X}^{2}_{ik}}}\,
\sqrt{\sum_{p=1}^{l}\mathbf{X}^{2}_{pj}},
\end{equation}
where $h$ is the hidden dimension and $l$ is the sequence length.

\textbf{Amber-Pruner:} Following~\citep{an2025amberprunerleveragingnm}, we (i) remove weight outliers outside the 0.5--99.5 percentiles, (ii) standardize the remaining weights, and (iii) score each activation as
$S_{\textit{Amb.-Pr.}}(\mathbf{X}_{ij})=|\mathbf{X}_{ij}|\cdot \mathcal{L}(\hat{\mathbf{W}}_{:,j})$,
where $\mathcal{L}(\cdot)$ is the channel-wise $\ell_2$ norm.

CLACT is context-aware (and for $l{=}1$ reduces to an $\ell_1$-type criterion), whereas Amber-Pruner leverages weight magnitudes but is not context-aware.

\subsection{Transformations for Error Mitigation}

\textbf{D-/S-/L-PTS} (dynamic/ static/ learnable per-token shift) centers activations near zero: $\hat{\mathbf{X}}=\mathbf{X}-\boldsymbol{\eta}$, and uses the compensated form
$\mathbf{Y}_{p}=((\hat{\mathbf{X}}\odot\mathbf{M})+\boldsymbol{\eta})\mathbf{W}^\top$~\citep{chua2024shiftbias}, where $\boldsymbol{\eta}=\overline{\mathbf{X}}$ for D-PTS; S-PTS uses a fixed $\boldsymbol{\eta}$ collected in a short warm-up; L-PTS learns $\boldsymbol{\eta}$.

\textbf{VAR} applies per-token variance correction after pruning:
$\mathbf{Y}_{p}=\boldsymbol{\nu}(\mathbf{X}\odot\mathbf{M})\mathbf{W}^\top$, with
\begin{equation}
\nu=\sqrt{\dfrac{\operatorname{Var}[\mathbf{X}]}{\operatorname{Var}[\mathbf{X}\odot\mathbf{M}]}}.
\label{eqn: variance_correction_parameter}
\end{equation}

\textbf{VAR+L-PTS} combines VAR scaling with the learnable shift. 
\textbf{R-Sparse} combines activation sparsity with a low-rank weight approximation~\citep{kamirul2025rsparsercnnsarship} (details in Appendix~\ref{sec:rsparse_details}).

For methods requiring calibration/learning, we use WikiText-2: S-PTS stores a fixed bias vector $\boldsymbol{\eta}$, L-PTS (and VAR+L-PTS) learns $\boldsymbol{\eta}$, and R-Sparse learns low-rank factors; all other methods require no learned parameters.

\subsection{Evaluation \& Models}
We evaluate the methods in Table~\ref{tab:strategies_descr} in two stages. First, we use \textbf{Core Datasets} (BoolQ, WinoGrande, PIQA, ARC-Easy) to screen all methods. We then focus on the most promising approaches on Qwen2.5-7B-Instruct and Llama3.1-8B-Instruct, and extend evaluation to \textbf{Extended Datasets} (HellaSwag, OpenBookQA, RTE, MMLU, Lambada\_standard, Lambada\_openai, IFEval). When calibration is required, we use WikiText-2, following common compression practice~\cite{egiazarian2024extreme,frantar2022gptq,van2024gptvq}. All results are obtained with LM Eval Harness~\cite{eval-harness}; dataset details are in Table~\ref{tab:dataset_descriptions}. We report results for Llama2-7B-chat, Llama3.1-8B-Instruct, Qwen2.5-7B-Instruct and Gemma3-4B-Instruct. For Qwen2.5-7B-Instruct, we do not sparsify key/query/value activations due to severe degradation observed in preliminary experiments.

% \subsection{Evaluation and Datasets}
% To analyze the efficiency of weight and activation pruning methods during the pretraining stage, we propose using the small Wikitext dataset. It will primarily serve for code debugging, pipeline validation, and addressing preliminary observations. For large-scale comprehensive studies and final evaluation of the developed methods, a 1B-token subset of the RedPajama~\cite{weber2024redpajama} dataset can be utilized.

% After that, finetuning of the pretrianed models is conducted with tulu-3-sft-mixture to enhance their capabilities for reasoning and question-answering tasks. Performance of the fine-tuned models can be estimated on accepted benchmarks, such as  MMLU~\cite{hendrycks2020mmlu}, BoolQ~\cite{clark2019boolq}, HellaSwag~\cite{zellers2019hellaswag}, WinoGrande~\cite{sakaguchi2021winogrande}, PiQA~\cite{tata2003piqa}, ARC-easy, and ARC-challenge~\cite{clark2018think}. The choice of baselines is similar to those in previous studies \cite{egiazarian2024extreme,frantar2022gptq,van2024gptvq}. 

\begin{figure}[ht]
    \centering

        \centering
         \includegraphics[width=\linewidth]{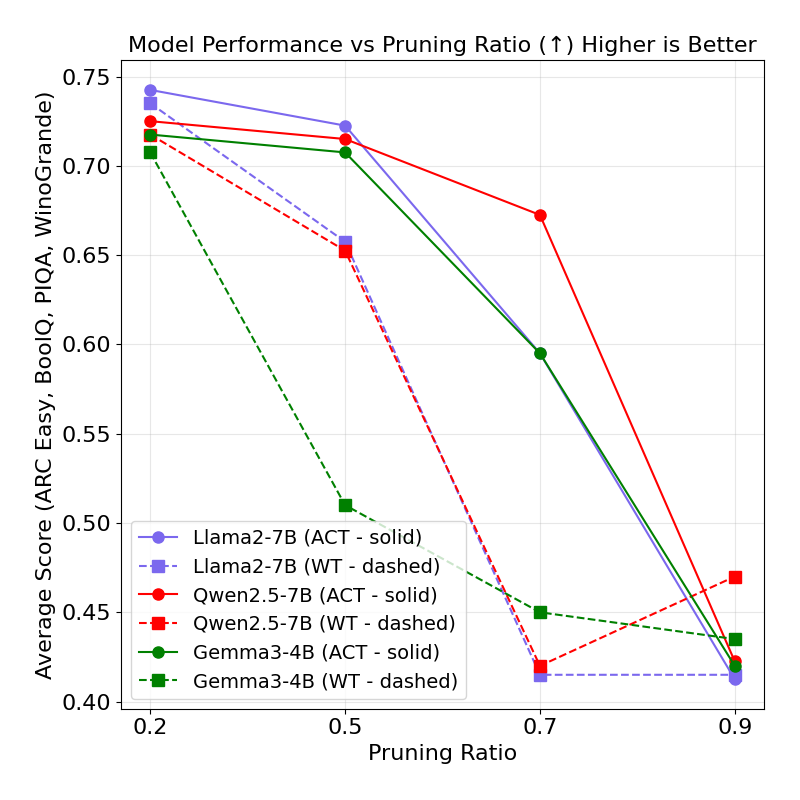}
        \caption{Comparison of \textbf{unstructured} \textbf{sparsity} in \textbf{activations} (\textbf{ACT}) and \textbf{weights} (\textbf{WT})  averaged across four datasets at varying sparsity ratios. \textbf{Higher is Better.} More detailed results are presented in Appendix Table~\ref{tab:unstructured_pruning}.}
    \label{fig:enter-label}

\end{figure}

\begin{figure}[ht]
        \centering
        \includegraphics[width=\linewidth]{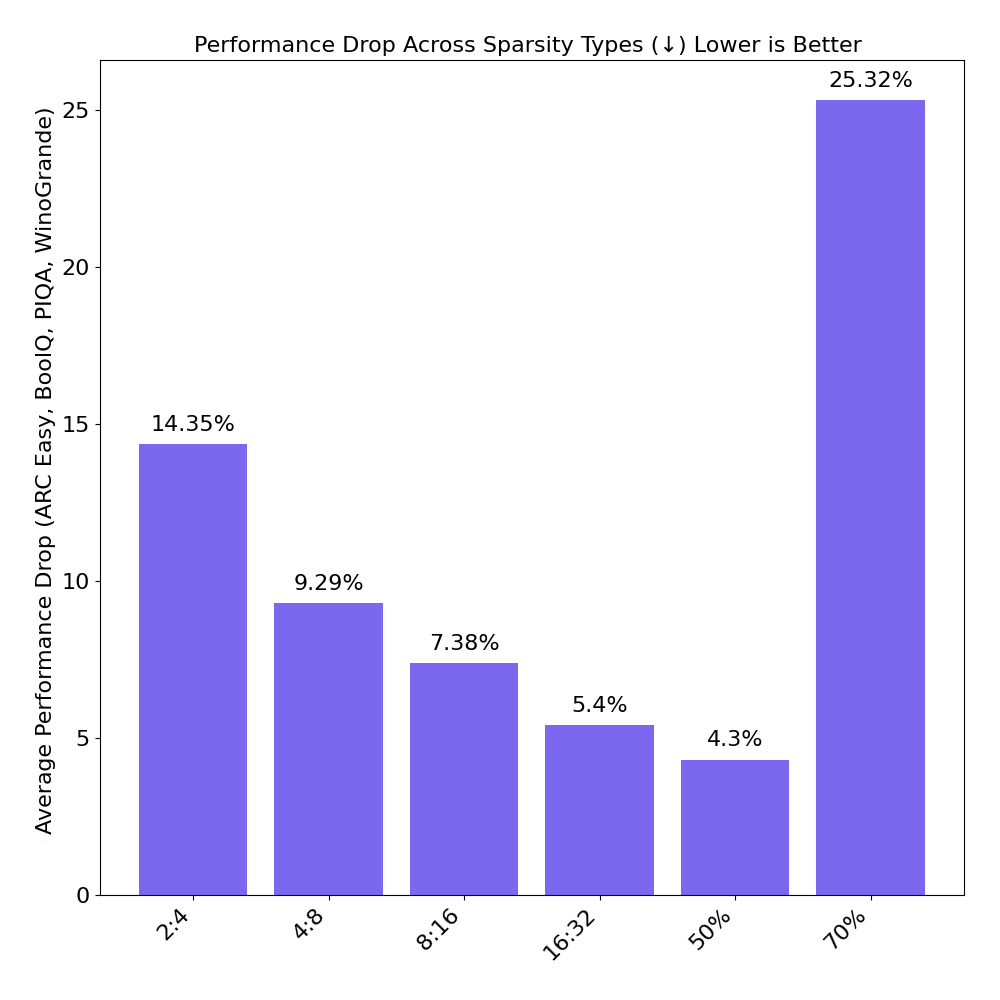}
        \caption{Comparison of sparsity patterns with unstructured sparsity. 50\% and 70\% correspond to unstructured sparsity.  \textbf{Lower is Better.} More detailed results are presented in Appendix Table~\ref{tab:sparsity_patterns}.}
        \label{fig:sparsity_patterns}
    
\end{figure}

\vspace{-0.5em}
\section{Results}
\label{sec:results}
\subsection{Sparse Weights vs. Activations}
In Figure~\ref{fig:enter-label} and Table~\ref{tab:unstructured_pruning}, we demonstrate that unstructured \textbf{weight sparsification causes greater model degradation} than unstructured activation sparsification at the same sparsity levels: \{20\%, 50\%, 70\%, 90\%\}.   For this evaluation, we specifically use unstructured magnitude-based sparsification, as it is less damaging than semi-structured sparsification and thus provides a lower bound on performance degradation. 
\subsection{Optimal semi-structured sparsity patterns}
Our preliminary investigation demonstrates that while the 16:32 pattern achieves performance close to unstructured sparsity (a 5.4\% drop versus 4.5\% for 50\% unstructured), it requires more metadata and greater resources for gather operations, as discussed in Section~\ref{sec:introduction}. Therefore, we focus on the 8:16 pattern, despite its higher performance drop of 7.38\%. For comparison, the 2:4 pattern results in a 14.35\% drop. These results are shown in Figure~\ref{fig:sparsity_patterns} and Table~\ref{tab:sparsity_patterns} in Appendix, we used only magnitude pruning to obtain these results. By demonstrating the superior model quality of 8:16 sparsity, our work incentivizes hardware designers to invest in or at least consider supporting the 8:16 pattern.

\begin{table}[th]
\centering
\caption{Average relative performance drop (\%, lower is better; negative = improvement) across four datasets, averaged over Llama2-7B-chat, Llama3.1-8B-Instruct, Qwen2.5-7B-Instruct and Gemma3-4B-Instruct. Act/Wt: activation/weight sparsity. Selection: ACT/CLACT/Amber-Pruner; transformations: VAR, D-PTS, S-PTS, L-PTS, R-SPARSE.  Methods marked with an asterisk (*) are proposed in this paper. Full results: Appendix~\ref{tab:main_two_four} and ~\ref{tab:main_eight_sixteen}.}
\label{tab:pruning_results}
\begingroup
\footnotesize
\setlength{\tabcolsep}{4.5pt}
\renewcommand{\arraystretch}{0.92}
\sisetup{table-format=+2.2, table-number-alignment=center}

\begin{tabular}{@{} l l l S @{}}
\toprule
\textbf{Target} & \textbf{Pattern} & \textbf{Method} & {\textbf{Avg drop ($\downarrow$)}} \\
\midrule
Act & u50  & ACT          & 3.82 \\

\addlinespace[2pt]
Wt  & 2{:}4  & WT           & 24.49 \\
Act & 2{:}4  & ACT          & 9.67 \\
Act & 2{:}4  & CLACT*       & \bfseries\tablenum{7.79} \\
Act & 2{:}4  & Amber-Pruner & 7.85 \\
Act & 2{:}4  & VAR*         & 6.09 \\
Act & 2{:}4  & D-PTS*       & 5.84 \\
Act & 2{:}4  & S-PTS        & \bfseries\tablenum{4.29} \\
Act & 2{:}4  & L-PTS*       & 8.79 \\
Act & 2{:}4  & R-SPARSE (64)  & 7.70 \\
Act & 2{:}4  & R-SPARSE (128) & 8.05 \\

\addlinespace[2pt]
Wt  & 8{:}16 & WT           & 17.68 \\
Act & 8{:}16 & ACT          & 5.47 \\
Act & 8{:}16 & CLACT*       & 2.29 \\
Act & 8{:}16 & Amber-Pruner & \bfseries\tablenum{1.56} \\
Act & 8{:}16 & VAR*         & 3.30 \\
Act & 8{:}16 & D-PTS*       & 2.07 \\
Act & 8{:}16 & S-PTS        & \bfseries\tablenum{0.61} \\
Act & 8{:}16 & L-PTS*       & 5.32 \\
Act & 8{:}16 & R-SPARSE (64)  & 1.52 \\
Act & 8{:}16 & R-SPARSE (128) & 2.63 \\
\bottomrule
\end{tabular}
\endgroup
\end{table}

\subsection{Results on Single/Multi-choice Datasets}
\subsubsection{Evaluation of Pruning Criteria}
We evaluate CLACT, Amber-Pruner, and magnitude pruning as a baseline. The main results for the 2:4 and 8:16 sparsity patterns are presented in Table~\ref{tab:pruning_results}.  On average, both CLACT and Amber-Pruner outperform magnitude pruning by at least 2\%, however, we observe no clear winner between them. As noted in Section~\ref{sec:pruning_criteria}, these criteria are designed for different purposes: CLACT adjusts based on context, while Amber-Pruner adjusts based on weight magnitudes. Notably, for Llama3.1-8B-Instruct under the 2:4 sparsity pattern, simple magnitude pruning outperforms both advanced criteria, underscoring model and architecture-specific sensitivities to pruning strategies. 

\subsubsection{Evaluation of Transformations}
Our main results are presented in Table~\ref{tab:pruning_results}. Surprisingly, we find that simple methods such as dynamic and static per-token shifts (D-PTS, S-PTS) outperform most other approaches.  The second most effective methods are VAR and R-SPARSE. We also observe that increasing the number of dimensions in R-SPARSE (from 64 to 128) leads to worse performance, which may indicate overfitting on the calibration data.  Finally, we note that L-PTS, the approach with learnable per-token shifts,  significantly underperforms compared to its static counterpart, S-PTS. 

\vspace{-0.5em}
\subsection{Instruction-Following Tasks}
Table~\ref{tab:spts_dpts_rsparse_var_ps_pl} presents instruction-following performance on the IFEval benchmark for Llama3.1-8B-Instruct and Qwen2.5-7B-Instruct, evaluated under two semi-structured sparsity patterns (2:4 and 8:16) and four activation transformation methods: S-PTS, D-PTS, R-Sparse, and VAR. First of all, we observe a strong model degradation on generative tasks. Second of all, we see that VAR is the strongest performer overall, especially for Llama3.1-8B-Instruct. S-PTS/D-PTS are competitive and lightweight, and R-Sparse lags significantly, particularly at 2:4.  We speculate that while semi-structured patterns are good for prefill stage in LLMs they significantly  degrade performance during decode stage. However, as discussed in Section~\ref{sec:introduction} decode stage for single vector can be accelerated with more flexible approaches.

\begin{table}[t]
\centering
\caption{Instruction-following (IFEval) prompt-level accuracy reported as PS/PL (prompt-level strict/loose accuracy).}
\label{tab:spts_dpts_rsparse_var_ps_pl}

\begingroup
\footnotesize
\setlength{\tabcolsep}{2 pt}     % was 4pt
\renewcommand{\arraystretch}{0.95}

\begin{tabular}{@{}l@{\hspace{2pt}}l@{\hspace{4pt}}cc@{}}
\toprule
\textbf{Model} & \textbf{Method} & \textbf{2{:}4} & \textbf{8{:}16} \\
\midrule
\multirow{4}{*}{\textbf{Llama3.1-8B}}
& ORIG     & 0.4455/0.4861 & 0.4455/0.4861 \\
& S-PTS    & 0.1682/0.1904 & 0.2995/0.3327 \\
& D-PTS    & 0.1941/0.2015 & 0.2828/0.3198 \\
& R-Sparse & 0.0869/0.0979 & 0.2089/0.2311 \\
& VAR      & \textbf{0.2237/0.2458} & \textbf{0.3161/0.3586} \\
\addlinespace[2pt]
\multirow{4}{*}{\textbf{Qwen2.5-7B}}
& ORIG     & 0.7135/0.7394 & 0.7135/0.7394 \\
& S-PTS    & 0.4325/0.5176 & 0.5194/0.5804 \\
& D-PTS    & 0.4399/0.5120 & \textbf{0.5434/0.5989} \\
& R-Sparse & 0.2736/0.3457 & 0.3697/0.4196 \\
& VAR      & \textbf{0.4565/0.5342} & 0.5249/0.5896 \\
\bottomrule
\end{tabular}
\endgroup
\end{table}
\vspace{-0.5em}

\subsection{Results for Unstructured Pruning}
We evaluate D-PTS, VAR, and two selection criteria in this experiment: CLACT and Amber-Pruner using the Llama3.1-8B-Instruct model. Results in Table~\ref{tab:unstruct} indicate that VAR is the most effective transformation under unstructured sparsity. Moreover, CLACT outperforms Amber-Pruner by a wider margin here than in our semi-structured pruning experiments.  These findings suggest two key insights: (1) No single method emerges as optimal for both unstructured and semi-structured sparsity. (2) The methods proposed in this work, VAR and CLACT, demonstrate strong generalization and are well-suited for both semi-structured and unstructured activation pruning. 

\begin{table}[t]
\centering
\caption{Performance comparison of pruning methods under 50\% and 70\% unstructured sparsity for Llama3.1-8B-Instruct model.}
\label{tab:unstruct}

\begingroup
\scriptsize
\setlength{\tabcolsep}{2.6pt} % tighter
\renewcommand{\arraystretch}{0.90}
\sisetup{table-number-alignment=center}

\resizebox{\columnwidth}{!}{%
\begin{tabular}{@{}l
S[table-format=1.3]
S[table-format=1.3]
S[table-format=1.3]
S[table-format=1.3]
S[table-format=2.3]@{}}
\toprule
\textbf{Method} 
& \multicolumn{1}{c}{\textbf{ArcE}}
& \multicolumn{1}{c}{\textbf{BoolQ}}
& \multicolumn{1}{c}{\textbf{PIQA}}
& \multicolumn{1}{c}{\shortstack{\textbf{Wino}\\\textbf{Grande}}}
& \multicolumn{1}{c}{\shortstack{\textbf{(\%) Avg.}\\\textbf{Drop}}} \\
\midrule

\textbf{Original} & 0.821 & 0.839 & 0.800 & 0.734 & \multicolumn{1}{c}{---} \\
\midrule
\multicolumn{6}{@{}c@{}}{\textbf{Unstructured 50\%}}\\[-1pt]
ACT          & 0.777 & 0.820 & 0.771 & 0.686 &  4.450 \\
D-PTS        & 0.786 & 0.825 & 0.781 & 0.690 &  3.600 \\
VAR          & 0.784 & 0.819 & 0.776 & 0.705 &  \textbf{3.470} \\
CLACT        & 0.780 & 0.825 & 0.766 & 0.700 &  3.890 \\
Amber & 0.768 & 0.820 & 0.763 & 0.702 &  4.450 \\
\midrule
\multicolumn{6}{@{}c@{}}{\textbf{Unstructured 70\%}}\\[-1pt]
ACT          & 0.558 & 0.631 & 0.647 & 0.548 & 25.320 \\
D-PTS        & 0.565 & 0.624 & 0.651 & 0.534 & 25.680 \\
VAR          & 0.614 & 0.651 & 0.676 & 0.532 & \textbf{22.660} \\
CLACT        & 0.555 & 0.604 & 0.627 & 0.524 & 27.670 \\
Amber & 0.487 & 0.594 & 0.590 & 0.539 & 30.680 \\
\bottomrule
\end{tabular}%
}
\endgroup
\end{table}

\subsection{Combination of Methods}
Next, we evaluated combinations of multiple approaches to explore potential performance gains. These combinations and their results are presented in Table~\ref{tab:paired_pruning_results}.   As shown, none of the evaluated combinations outperforms any single method, highlighting the challenges of naively combining them. 

\begin{table}[t]
\centering
\caption{Llama3.1-8B-Instruct with 8{:}16 activation sparsity: aggregated results across layer subsets. \textbf{ORIG. AVG.} = 0.6726. Drop is computed w/o perplexity. Full results: Appendix~\ref{tab:ls_lpts_var_by_layers}.}
\label{tab:l3_8b_816_summary}

\begingroup
\scriptsize
\setlength{\tabcolsep}{2.6pt}
\renewcommand{\arraystretch}{0.90}

\resizebox{\columnwidth}{!}{%
\begin{tabular}{@{}l l r r r@{}}
\toprule
\textbf{Method} & \textbf{Layers} & \textbf{PPL} & \textbf{AVG.} & \textbf{Drop $\downarrow$} \\
\midrule
LS+L-PTS       & all                 & 9.6036 & 0.6047 & 10.90\% \\
LS+L-PTS       & key,out,gate,down    & 8.3483 & 0.6385 &  5.43\% \\
LS+L-PTS       & key,value,gate,down  & 8.0821 & 0.6503 & \textbf{3.56}\% \\
\addlinespace[2pt]
LS+L-PTS+VAR   & all                 & 9.4983 & 0.6056 & 10.60\% \\
LS+L-PTS+VAR   & key,out,gate,down    & 8.2930 & 0.6422 &  4.64\% \\
LS+L-PTS+VAR   & key,value,gate,down  & 8.0259 & 0.6516 & \textbf{3.36}\% \\
\bottomrule
\end{tabular}%
}
\endgroup
\end{table}

\subsection{Layer Sensitivity}
\label{sec:layer_sensitivity}
We study layer sensitivity to 8:16 activation sparsity on Llama3.1-8B-Instruct using the Extended Datasets (Table~\ref{tab:l3_8b_816_summary}). This analysis focuses on learnable methods, mainly L-PTS and LS (learnable diagonal scaling). Although learnable approaches underperform on average in our broader experiments, the best 8:16 configurations for Llama3.1-8B-Instruct rely on learnable parameters (Table~\ref{tab:l3_8b_816_summary}). We observe that the FFN up projection and the attention out projection\footnote{The output projection of the attention mechanism. It combines outputs from all attention heads and projects them back to the model’s hidden dimension.} are the most sensitive: pruning them causes the largest drops, suggesting they should be preserved or handled with extra care. While this may not generalize to all layers, it indicates that layer importance under activation sparsity is highly non-uniform.

\subsection{Analysis of Qwen2.5-7B-Instruct Anomalous Improvements}
\label{sec:qwen-anomaly}
 
Several configurations for Qwen2.5-7B-Instruct under 8:16 sparsity outperform dense baseline on certain benchmarks (e.g., D-PTS: $-8.28\%$, R-SPARSE(64): $-6.90\%$. Unlike the other three models, Qwen2.5-7B-Instruct required excluding key/query/value projections from sparsification due to severe degradation observed in preliminary experiments. This means that only a subset of linear layers is actually pruned, reducing the overall perturbation applied to the network. With fewer layers affected, the risk of cascading errors is lower, and the remaining pruned layers may benefit from an implicit regularization effect without being offset by damage elsewhere.
 
\vspace{-0.5em}
\section{Discussion}
The performance gap between multiple/single-choice benchmarks (e.g., BoolQ, PIQA) and IFEval likely stems from differences in the inference stages they emphasize. Core QA benchmarks primarily stress the prefill phase, whereas IFEval evaluates both prefill and autoregressive generation.  Our evaluation remains valid: semi-structured patterns like 2:4 and 8:16 are especially effective at accelerating the prefill stage, which often dominates inference latency. 

\subsection{Implications of IFEval Degradation for Generative Deployment}
\label{sec:ifeval-discussion}
 
While activation sparsity preserves multiple-choice QA performance remarkably well, our IFEval results  reveal a substantially different picture for generative, instruction-following tasks.
A ${\sim}26\%$ degradation in instruction adherence under 8:16 sparsity is likely unacceptable for many such use cases without additional mitigation. We hypothesize that the gap between QA and generative performance stems from how sparsity interacts with different inference stages.
Multiple-choice QA benchmarks primarily stress the \emph{prefill} phase, where the model processes the entire prompt in parallel and produces a single-token or few-token classification response.
Semi-structured sparsity patterns are well-suited to this regime because the sparsification is applied to large activation matrices with favorable statistics.
In contrast, IFEval evaluates both prefill and \emph{autoregressive generation}, where errors introduced by sparsification compound across hundreds of generated tokens. During the decode stage, each token's representation is a single vector, and block-structured sparsity patterns impose rigid constraints on which elements can be zeroed. We emphasize that this limitation is not unique to our approach; the performance gap between generative and QA tasks under compression is well-documented across both sparsification and quantization methods~\citep{ding2026llms}.

\vspace{-0.5em}
\section{Conclusion}
This work establishes that post-training activation pruning is significantly more accuracy-preserving than weight pruning in large language models. Across four diverse architectures (Llama2-7B, Llama3.1-8B, Qwen2.5-7B, and Gemma3-4B), we demonstrate that activation sparsity consistently retains model capabilities better than weight sparsity at matched sparsity levels.

Our evaluation reveals that lightweight error mitigation techniques particularly CLACT, D-PTS, and VAR establish strong, hardware friendly baselines requiring minimal calibration data. Through systematic analysis of semi-structured patterns, we find that 16:32 approaches unstructured 50\% sparsity in fidelity, while 8:16 emerges as the optimal near term target.

% Despite limitations in hardware emulation and generative task evaluation, these findings provide concrete evidence that expanding hardware support beyond static 2:4 weight sparsity can unlock significant efficiency gains without compromising model quality. We urge hardware designers to prioritize native support for flexible N:M activation sparsity patterns, particularly 8:16, which represents a practical stepping stone toward truly adaptive, input-responsive acceleration of large language models. This direction bridges the critical gap between theoretical sparsity benefits and real-world inference efficiency, potentially transforming how we deploy large language models at scale.

% In summary, this work not only delivers practical, high performing methods for post-training activation sparsity but also provides compelling, empirically grounded motivation for hardware architects to support 8:16 structured sparsity for activations. 
\vspace{-0.5em}
\section{Limitations}
Key limitations: First, all evaluations use software emulation without hardware measurements of speedup or energy efficiency. Second, layer sensitivity analysis remains preliminary; while FFN up-projections and attention out-projections appear most vulnerable, broader architectural studies are needed. Third, generative performance (IFEval) degrades significantly more than multiple-choice QA under sparsity, revealing an evaluation bias toward prefill-dominated workloads. The anomalous improvements on Qwen2.5-7B-Instruct for some benchmarks further highlight dataset-specific artifacts rather than genuine capability preservation. Our hardware overhead analysis in Appendix~\ref{app:hardware} is a rough estimate and more precise analysis is required.

\vspace{0.5em}
\small \textbf{Acknowledgment}: The work was supported by the grant for research centers in the field of AI provided by the Ministry of Economic Development of the Russian Federation in accordance with the agreement 000000C313925P4F0002 and the agreement with Applied AI Institute №139-10-2025-033.
% \paragraph{Acknowledgment on LLM assisted writing:}
% This paper used open access Qwen3-Max, in some parts of the paper,  for proofreading and text rephrasing in accordance with formal style.
\clearpage
\bibliography{bibs}

\clearpage
\appendix
\section*{Appendix}

\section{Hardware Implications and Computational Overhead Analysis}
\label{app:hardware}

While our empirical results demonstrate significant accuracy benefits from flexible N:M activation sparsity, the practical value of these techniques depends critically on whether hardware implementations can overcome the computational overhead of dynamic sparsification. This section provides a comprehensive break-even analysis to determine the hardware conditions required for activation sparsity to deliver net performance and efficiency gains.

% \subsection{Break-Even Thresholds for Activation Sparsity}
The fundamental challenge for activation sparsity is that the theoretical memory bandwidth reduction must overcome the overhead introduced by dynamic sparsification operations. Based on literature and performance models, we estimate break-even thresholds where benefits begin to outweigh costs.

\subsection{Energy-Delay Product Analysis}
\label{sec:edp}
The Energy-Delay Product (EDP) provides a comprehensive metric for evaluating whether activation sparsity delivers net efficiency benefits. For 8:16 sparsity to be worthwhile from an EDP perspective, it must overcome both computational overhead and energy costs of the sparsification process itself.

We model EDP improvement for semi-structured sparsity patterns as:
\[
\text{EDP}_{\text{improvement}} = \frac{\text{EDP}_{\text{dense}}}{\text{EDP}_{\text{sparse}}} \approx \frac{r \cdot \eta}{1 + \alpha}
\]

Where:
\begin{itemize}
    \item $r = 2.0$ is the theoretical bandwidth reduction ratio for 8:16 sparsity
    \item $\eta = 0.85$ is the hardware utilization efficiency (representative of practical implementations)
    \item $\alpha = 0.3$ is the overhead factor from sparsification operations
\end{itemize}

This overhead factor $\alpha$ is calibrated from real measurements~\cite{fang2024maskllm} demonstrated that dynamic activation sparsification incurs 30--35\% additional latency on current hardware without native support, broken down as:
\begin{itemize}
    \item Activation magnitude computation and block-wise selection
    \item Mask application and metadata handling
    \item  Error mitigation techniques (D-PTS, VAR), which is not included in~\cite{fang2024maskllm}.
\end{itemize}

Solving for the minimum hardware acceleration factor $k$ required for net EDP benefits:
\[
r \cdot \eta > k \cdot (1 + \alpha)
\]
\[
2.0 \cdot 0.85 > k \cdot (1 + 0.3)
\]
\[
k > \frac{1.7}{1.3} \approx 1.31
\]

However, due to our imprecise estimations we will consider a higher \textit{amortized} $k > 1.6\times$ required for speedup.

\subsection{Hardware Implementation Requirements}
To achieve the required $>$1.6$\times$ speedup and cross the break-even threshold, hardware must include:

\begin{itemize}
    \item \textbf{Dedicated sparsity controllers}: On-chip circuitry that can generate sparsity masks with minimal latency, reducing the 15--20\% selection overhead
    \item \textbf{Hardware-supported statistical units}: Specialized units for variance/mean calculation required by error mitigation techniques, eliminating their computational overhead
    \item \textbf{Hierarchical sparsity support}: Different patterns for different layer types based on sensitivity analysis (Section~\ref{sec:layer_sensitivity})
    \item \textbf{Bandwidth-optimized gather operations}: Specialized memory controllers that maintain high efficiency despite irregular access patterns
\end{itemize}

Recent hardware developments show promising directions: NVIDIA's Blackwell architecture includes a dedicated hardware decompression engine \cite{jarmusch2025microbenchmarking}, while the LazyGPU microarchitecture enables lazy memory request issuing \cite{liu2025sparsity}.

This analysis provides concrete targets for hardware designers: 8:16 activation sparsity can deliver significant accuracy preservation  while achieving net performance gains, but only if hardware can deliver $>$1.6$\times$ speedup for sparse operations and efficiently support the statistical computations required by error mitigation techniques. 

% Required packages (ensure these are in your preamble):
% \usepackage{booktabs}
% \usepackage{amsmath}
% \usepackage{array}

\subsection{Microarchitectural Implementation Costs \& Complexity Analysis}
\label{subsec:impl_costs}

Precise implementation cost estimates are inherently challenging to formalize, as commercial GPU/TPU vendors typically do not disclose detailed microarchitectural specifications; many of these design parameters are governed by strict NDAs. The figures presented here are therefore engineering estimates grounded in publicly available microarchitectural analyses. While the absolute index width increases, the control logic scales sub-linearly because the combinatorial encoder/decoder can be implemented via lightweight lookup tables and bit-packing circuits rather than full-width arithmetic units. Drawing on published sparse-accelerator design studies~\cite{lin2023efficient} and microarchitectural analyses of Blackwell-class decompression engines, we conservatively estimate that extending an existing 2:4 pipeline to support 8:16 will incur an incremental die area overhead of \textbf{$<2\%$}.  The 8:16 pattern incurs a \textbf{16.7\% higher metadata bandwidth} relative to 2:4 ($0.875 / 0.75 \approx 1.167$). To provide a structured overview of implementation trade-offs, we summarize the relative complexity across four key dimensions for 2:4 vs.\ 8:16 activation sparsity in Table~\ref{tab:complexity_comparison}. Ratings reflect incremental cost relative to a baseline dense tensor core.
We assess implementation complexity across four dimensions, including estimated Non-Recurring Engineering (NRE) cos the one-time design, validation, and integration effort required to extend a tensor core to support a new sparsity pattern, excluding per-unit manufacturing costs.
\begin{table*}[t]
    \centering
    \caption{Qualitative complexity comparison across four microarchitectural dimensions for 2:4 vs.\ 8:16 activation sparsity. NRE = Non-Recurring Engineering cost (one-time design/validation effort).}
    \label{tab:complexity_comparison}
    \resizebox{\textwidth}{!}{%
    \begin{tabular}{@{} l c c p{7.8cm} @{}}
        \toprule
        \textbf{Dimension} & \textbf{2:4} & \textbf{8:16} & \textbf{Justification \& References} \\
        \midrule
        \textbf{Metadata Overhead} & Low ($0.75$ bits/elt) & Low--Med ($0.875$ bits/elt) & Combinatorial encoding scales logarithmically; $16.7\%$ increase is marginal \\
        \addlinespace
        \textbf{Controller Logic} & Low (2-bit decoders) & Medium (14-bit unpacking) & Requires wider LUTs \& dynamic gather scheduling, but shares base sparse pipeline~\cite{lin2023efficient, fang2024maskllm} \\
        \addlinespace
        \textbf{Memory Bandwidth} & Low (halves fetches) & Low--Med (+$16.7\%$ metadata) & Net bandwidth drops due to $2\times$ activation pruning; metadata fits HBM3 headroom \\
        \addlinespace
        \textbf{NRE Cost Tier} & Low (mature IP) & Medium (index + gather opt.) & Validates dynamic mask generation without full tensor-core redesign~\cite{liu2025sparsity} \\
        \bottomrule
    \end{tabular}%
    }
\end{table*}

\section{R-Sparse Details}
\label{sec:rsparse_details}
Finally, we include \textbf{R-Sparse}~\citep{kamirul2025rsparsercnnsarship}, which combines activation sparsity with a low-rank approximation of the weight matrix. Instead of pruning solely by magnitude, R-Sparse decomposes the computation into two parts: (i) sparse channels with high-magnitude activations, and (ii) a low-rank component obtained via SVD of $\mathbf{W}$ that approximates the contribution of pruned activations. 

Formally, the linear layer
\begin{equation}
    \mathbf{Y} = \mathbf{X}\mathbf{W}^\top
\end{equation}
is approximated as
\begin{equation}
    \mathbf{Y} \approx \mathbf{Y}_s + \mathbf{Y}_r,
\end{equation}
where

 $$   \mathbf{Y}_s = \sigma_{t(s)}(\mathbf{X})\mathbf{W}^\top, $$ 
   
  $$  \mathbf{Y}_r = (\mathbf{X}-\sigma_{t(s)}(\mathbf{X}))(A_rB_r)^\top. $$

Here $\sigma_{t(s)}(\cdot)$ denotes sparsification of activations with threshold $t(s)$, and $A_rB_r^\top$ is the rank-$r$ approximation of $\mathbf{W}$ obtained from its truncated SVD. The trade-off between $\mathbf{Y}_s$ and $\mathbf{Y}_r$ is determined by a sparsity budget $s$ and rank $r$, which can be optimized via evolutionary search.

\section{Main extended results}
\label{sec:main_extended}
Here, we present un-aggregated results. Comparisons between different sparsity patterns is presented in Table~\ref{tab:sparsity_patterns}. In Table~\ref{tab:main_two_four} and Table~\ref{tab:main_eight_sixteen} we present the results of the error mitigation strategies and the selection criteria. Table~\ref{tab:main_two_four} reports the results of the combined methods.  Table~\ref{tab:unstructured_pruning} compares unstructured and semi-structured sparsity. Finally, in Table~\ref{tab:ls_lpts_var_by_layers} we demonstrate results when some of the layers are excluded to evaluate layer sensitivity.

\begin{table*}[htbp]
\caption{Performance comparison of different sparsity patterns on Llama3.1-8B-Intsruct across various benchmarks. Values represent accuracy scores, with the last column showing the average performance drop relative to the original model.}
\centering
%\resizebox{0.8\textwidth}{!}{%
\begin{tabular}{@{}l *{4}{S[table-format=1.4]} S[table-format=2.2\%] @{}}
\toprule
 & \textbf{ARC Easy} & \textbf{BoolQ} & \textbf{PIQA} & \textbf{WinoGrande} & \textbf{Avg Drop ($\downarrow$)} \\
\midrule
\textbf{Original} & 0.8207 & 0.8391 & 0.8003 & 0.7340 & \\
\midrule
2:4 & 0.6837 & 0.7261 & 0.7163 & 0.6110 & 14.35\% \\
4:8 & 0.7272 & 0.7810 & 0.7529 & 0.6393 & 9.29\% \\
8:16 & 0.7525 & 0.7969 & 0.7568 & 0.6551 & 7.38\% \\
16:32 & 0.7698 & 0.8082 & 0.7688 & 0.6771 & 5.40\% \\
50\% unstructured & 0.7820 & 0.8198 & 0.7714 & 0.6858 & 4.30\% \\
70\% unstructured & 0.5580 & 0.6311 & 0.6474 & 0.5477 & 25.32\% \\
\bottomrule
\end{tabular}%
\label{tab:sparsity_patterns}
\end{table*}

\begin{table*}[hbt]
\centering
\caption{\textbf{A comparison of combined approaches with 8:16 semi-structured sparsity.} Average relative performance (\%) across four datasets. Values indicate performance drops (lower is better), negative values signify performance improvement. 
Full, non-aggregated results are available in Appendix~\ref{tab:main_eight_sixteen}.}
\label{tab:paired_pruning_results}
\resizebox{\textwidth}{!}{%
\begin{tabular}{@{}l *{4}{S[table-format=2.2\%]} S[table-format=2.2\%] @{}}
\toprule
 & \multicolumn{4}{c}{\textbf{Models}} & \\
\cmidrule(lr){2-5}
\textbf{Method} & \textbf{Llama2-7B-chat} & \textbf{Qwen2.5-7B-Instruct} & \textbf{Gemma3-4B-Instruct} & \textbf{Llama3.1-8B-Instruct} & \textbf{Average Drop ($\downarrow)$} \\
CLACT + PTS & 5.63\% & -5.06\% & 0.50\% & 8.55\% & \textbf{2.40}\% \\
CLACT + VAR & 5.07\% & -2.90\% & 0.54\% & 8.59\% & 2.82\% \\
Amber-Pruner + PTS & 6.16\% & -3.47\% & 0.17\% & 7.42\% & 2.57\% \\
Amber-Pruner + VAR & 4.74\% & -3.63\% & -0.16\% & 8.39\% & \textbf{2.34}\% \\
L-PTS + VAR & 6.87\% & 2.86\% & 3.41\% & 7.15\% & 5.07\% \\ 
\bottomrule
\end{tabular}%
}
\end{table*}

\section{Datasets}
\label{sec:datasets}
Detailed description of the datasets is given in Table~\ref{tab:dataset_descriptions}.

\begin{table*}[ht]
\centering
\caption{Datasets used to evaluate hypotheses. \textit{Prompt-level strict accuracy} is the fraction of prompts for which all verifiable instructions in the prompt are followed exactly as stated. \textit{Instruction-level strict accuracy} is the fraction of individual instructions that are followed exactly as stated, averaged across all instructions.}
\label{tab:dataset_descriptions}

\begingroup
\small
\setlength{\tabcolsep}{3pt}
\renewcommand{\arraystretch}{0.9}
\begin{tabular}{p{0.2\textwidth} p{0.63\textwidth} p{0.1\textwidth}}
\toprule
\textbf{Dataset} & \textbf{Description} & \textbf{Metric} \\
\midrule

WikiText-2 \citep{merity2016pointer} &
A collection of over 100 million tokens extracted from the set of verified Good and Featured articles on Wikipedia. &
Perplexity \\

\midrule

ARC-Easy \citep{arc} &
QA benchmark for genuine grade-school level, multiple-choice science questions. The dataset contains 2251 examples for training, 570 for development and 2376 for testing. &
Accuracy \\

\midrule

ARC\_Challenge \citep{arc} &
QA benchmark for more difficult grade-school level science questions, part of the AI2 Reasoning Challenge. Designed to require deeper reasoning than ARC-Easy. &
Accuracy \\

\midrule

BoolQ \citep{clark2019boolq} &
QA benchmark for yes/no questions. The dataset contains 9427 examples for training and 3270 for testing. &
Accuracy \\

\midrule

PIQA \citep{bisk2020piqa} &
Physical commonsense QA benchmark for choosing the right answer between two options. Contains 16K train, 2K dev, and 3K test examples. &
Accuracy \\

\midrule

WinoGrande \citep{sakaguchi2021winogrande} &
QA benchmark for pronoun resolution with adversarial filtering. Contains 40K train, 1267 dev, and 1767 test examples. &
Accuracy \\

\midrule

HellaSwag \citep{zellers2019hellaswag} &
Commonsense reasoning benchmark for sentence completion, designed to be easy for humans but hard for models. Contains 70K train and 10K validation examples. &
Accuracy \\

\midrule

OpenBookQA \citep{mihaylov2018openbookqa} &
Open-book question answering dataset requiring retrieval of elementary science facts. Contains 5957 4-way multiple-choice questions. &
Accuracy \\

\midrule

RTE \citep{dagan2005pascal, bar2006second} &
Recognizing Textual Entailment datasets from PASCAL challenges. Task is to classify if a hypothesis is entailed by a premise. &
Accuracy \\

\midrule

MMLU \citep{hendrycks2020measuring} &
Massive Multitask Language Understanding benchmark covering 57 subjects across STEM, humanities, and social sciences. Measures multitask accuracy. &
Accuracy \\

\midrule

Lambada\_Standard \citep{paperno2016lambada} &
Word prediction task requiring broad discourse context. Target word is unpredictable from local context alone. &
Accuracy \\

\midrule

Lambada\_OpenAI \citep{paperno2016lambada} &
LAMBADA test set preprocessed by OpenAI for standardized evaluation. Task remains final word prediction with long-range dependencies. &
Accuracy \\

\midrule

GSM8K \citep{cobbe2021gsm8k} &
Grade school math word problems requiring multi-step reasoning. Contains 7.5K train and 1.3K test examples. &
\begin{tabular}[c]{@{}c@{}}Accuracy \\ (Strict) \\ Accuracy \\ (Flexible)\end{tabular} \\

\midrule

IFEval \citep{zhou2023instructionfollowingevaluationlargelanguage} &
Benchmark with 541 prompts containing verifiable instructions to measure instruction-following fidelity. &
\begin{tabular}[c]{@{}c@{}}Accuracy \\ (Prompt-level) \\ Accuracy \\ (Instruct-level)\end{tabular} \\

\bottomrule
\end{tabular}
\endgroup
\end{table*}

\begin{table*}[hbt]
\centering
\caption{The performance of models with applied unstructured activation pruning. We show that even with severe sparsity (70-90\%) models were able to perform decently on our benchmarks. \textbf{ACT} stands for activations pruning,  \textbf{WT} — for weight pruning.  \textbf{OUT} denotes values more than $10^3$, according accuracy scores most likely correspond to random.}

\scalebox{0.88}{\begin{tabular}{lcccccc}
\toprule
\textbf{Pruning} & \textbf{WikiText2} $\downarrow$ & \textbf{ARC Easy} & \textbf{BoolQ} & \textbf{PIQA} & \textbf{WinoGrande} & \textbf{Drop}~$(\downarrow) \%$\\
\midrule
\multicolumn{6}{l}{\textbf{Llama2-7B-chat}} \\
\midrule
\quad \textbf{Base} & 6.94 & 0.74 & 0.80 & 0.76 & 0.66 & -\\
\midrule
\quad 0.2 \textbf{ACT} & 6.96 & 0.74 & 0.80 & 0.77 & 0.66 & $\textbf{-0.33}\%$\\
\quad 0.2 \textbf{WT} & 7.49 & 0.72 & 0.80 & 0.76 & 0.66 & $0.68\%$ \\
\hline
\quad 0.5 \textbf{ACT} & 7.53 & 0.70 & 0.78 & 0.75 & 0.66 & $\textbf{2.32}\%$  \\
\quad 0.5 \textbf{WT} & 18.72 & 0.60 & 0.72 & 0.70 & 0.61  & $11.10\%$ \\
\hline
\quad 0.7 \textbf{ACT} & 20.11 & 0.56 & 0.64 & 0.65 & 0.53  & $\textbf{19.62}\%$ \\
\quad 0.7 \textbf{WT} & \textbf{OUT} & 0.27 & 0.38 & 0.54 & 0.47  & $43.44\%$ \\
\hline
\quad 0.9 \textbf{ACT} & \textbf{OUT} & 0.26 & 0.38 & 0.52 & 0.49  & $43.39\%$ \\
\quad 0.9 \textbf{WT} & \textbf{OUT} & 0.27 & 0.38 & 0.53 & 0.48  & $43.39\%$ \\
\midrule
\multicolumn{6}{l}{\textbf{Qwen2.5-7B-Instruct}} \\
\midrule
\quad \textbf{Base} & 7.46 & 0.69 & 0.86 & 0.75 & 0.60  & - \\
\midrule
\quad 0.2 \textbf{ACT} & 7.48 & 0.69 & 0.86 & 0.74 & 0.61  & $\textbf{2.37}\%$ \\
\quad 0.2 \textbf{WT} & 8.03 & 0.67 & 0.86 & 0.74 & 0.60  & $3.42\%$ \\
\hline
\quad 0.5 \textbf{ACT} & 8.3 & 0.67 & 0.87 & 0.74 & 0.58  & $3.87\%$ \\
\quad 0.5 \textbf{WT} & 43.6 & 0.56 & 0.80 & 0.68 & 0.57  & $\textbf{3.42}\%$ \\
\hline
\quad 0.7 \textbf{ACT} & 18.7 & 0.6 & 0.81 & 0.70 & 0.58  & $\textbf{3.87}\%$ \\
\quad 0.7 \textbf{WT} & \textbf{OUT} & 0.28 & 0.38 & 0.54 & 0.48  & $12.12\%$ \\
\hline
\quad 0.9 \textbf{ACT} & \textbf{OUT} & 0.25 & 0.38 & 0.54 & 0.52  & $44.22\%$ \\
\quad 0.9 \textbf{WT} & \textbf{OUT} & 0.25 & 0.58 & 0.54 & 0.51  & $\textbf{36.35}\%$ \\
\midrule
\multicolumn{6}{l}{\textbf{Gemma3-4B-Instruct}} \\
\midrule
\quad \textbf{Base} & 17.29 & 0.72 & 0.84 & 0.72 & 0.62 & -\\
\midrule
\quad 0.2 \textbf{ACT} & 17.60 & 0.71 & 0.84 & 0.72 & 0.60  & $\textbf{3.35}\%$ \\
\quad 0.2 \textbf{WT} & 18.93 & 0.68 & 0.84 & 0.72 & 0.59  & $4.74\%$ \\
\hline
\quad 0.5 \textbf{ACT} & 22.39 & 0.71 & 0.83 & 0.72 & 0.57  & $\textbf{4.80}\%$ \\
\quad 0.5 \textbf{WT} & 273 & 0.36 & 0.55 & 0.61 & 0.52  & $30.89\%$ \\
\hline
\quad 0.7 \textbf{ACT} & 88 & 0.55 & 0.63 & 0.66 & 0.54  & $\textbf{19.57}\%$ \\
\quad 0.7 \textbf{WT} & \textbf{OUT} & 0.27 & 0.49 & 0.53 & 0.51  & $38.81\%$ \\
\hline
\quad 0.9 \textbf{ACT} & \textbf{OUT} & 0.26 & 0.38 & 0.54 & 0.50  & $42.64\%$ \\
\quad 0.9 \textbf{WT} & \textbf{OUT} & 0.25 & 0.45 & 0.52 & 0.52  & $\textbf{40.60}\%$ \\
\bottomrule
\end{tabular}}
\label{tab:unstructured_pruning}
\end{table*}

\begin{table*}[hbt]
\centering
% \begin{minipage}[t]{0.49\textwidth}
\centering
\caption{\textbf{Semi-Structured 2:4 Sparsification} - performance Metrics, for calibration, when it is required, and perplexity we use WikiText2. Average Drop is computed without accounting for perplexity.}
\resizebox{\textwidth}{!}{%
\begin{tabular}{lcccccc}

\toprule
\textbf{Pruning} & \textbf{WikiText2} $\downarrow$ & \textbf{ARC Easy} & \textbf{BoolQ} & \textbf{PIQA} & \textbf{WinoGrande} & \textbf{Average Drop \%} \\
\midrule
\quad \textbf{Llama2-7B-chat}  & 6.94 & 0.74 & 0.80 & 0.76 & 0.66 &- \\
\midrule
 \textbf{ACT} & 10.23 & 0.66 & 0.71 & 0.71 & 0.60 & 9.43\%\\
 \textbf{WT} & 42.40 & \textbf{0.57} & 0.65 & \textbf{0.69} & 0.56 & 16.52\%\\
 \textbf{D-PTS} & 9.38 & 0.64 & 0.68 & 0.71 & 0.61 & 10.67\% \\
 \textbf{S-PTS} & 9.36 & 0.66 & 0.68 & 0.71 & 0.60 & 10.37\% \\
 \textbf{VAR} & \textbf{8.31} & 0.67 & 0.69 & 0.72 & 0.59 & 9.76\%\\
 \textbf{CLACT} & 8.23 & 0.65 & 0.72 & 0.71 & 0.63 & 8.32\% \\
 \textbf{Amber-Pruner} & 9.24 & 0.64 & 0.68 & \textbf{0.69} & 0.60 & 11.70\% \\
 \textbf{LPTS} & 8.89 & 0.65 & \textbf{0.60} & 0.72 & \textbf{0.59} & 13.13\% \\
 \textbf{LPTS + VAR} & 8.39 & 0.67 & 0.63 & 0.72 & 0.60 & 11.47\% \\

  \textbf{R-SPARSE (64)} &9.19&0.66&0.63&0.69&0.59&12.90\% \\
   \textbf{R-SPARSE (128)} & 9.29&0.65&0.65&	0.70&0.59&12.23\%		\\

\midrule
\quad \textbf{Llama3.1-8B-Instruct} & 7.21  & 0.82 & 0.84 & 0.80 & 0.73 & - \\
\midrule
 \textbf{ACT} & 16.61 & 0.68 & 0.73 & 0.72 & 0.61 & 14.35\%\\
 \textbf{WT} & 20.14 & 0.41 & 0.57 & 0.60 & 0.54 & 33.63\%\\
 \textbf{PTS} & 16.4 & 0.69 & 0.73 & 0.72 & 0.60 & 14.59\%\\
 \textbf{S-PTS (N-100)} & 16.5 & 0.67 & 0.74 & 0.72 & 0.60 & 14.61\%\\
 \textbf{S-PTS (N-200)} & 16.5 & 0.68 & 0.73 & 0.72 & 0.61 & 14.31\%\\
 \textbf{VAR} & 14.17 & 0.70 & 0.73 & 0.73 & 0.62 & 13.11\%\\
 \textbf{CLACT} & 19.49 & 0.65 & 0.71 & 0.69 & 0.59 & 17.27\%\\
 \textbf{WANDA} & 15.86 & 0.66 & 0.74 & 0.69 & 0.61 & 15.01\%\\
 \textbf{L-PTS} & 12.77 & 0.71 & 0.71 & 0.73 & 0.59 & 14.13\%\\
 \textbf{L-PTS + VAR} & 12.40 & 0.73 & 0.71 & 0.73 & 0.60 & 13.49\%\\

\textbf{R-SPARSE (64)} & 15.07  &0.69&0.72&0.71&0.61&15.28\% \\
\textbf{R-SPARSE (128)} & 16.09 &0.67&0.71&0.70&0.61&16.34\% \\

\midrule
%\multicolumn{6}{l}{} \\
\quad \textbf{Qwen2.5-7B-Instruct} & 7.46 & 0.69 & 0.86 & 0.75 & 0.60 & - \\
\midrule
 \textbf{ACT}  & 10.06 & 0.65 & 0.86 & 0.72 & 0.54 & 4.95\% \\
 \textbf{WT} & 35.37 & 0.53 & 0.78 & 0.68 & 0.54 & 12.96\%  \\
 \textbf{D-PTS} & 10.07 & 0.79 & 0.86 & 0.76 & 0.66 & -6.46\% \\
 \textbf{S-PTS} & 10.74 & 0.78 & 0.84 & 0.74 & 0.65 & -4.43\% \\
 \textbf{VAR} & 13.95 & 0.74 & 0.83 & 0.74 & 0.61  & -1.48\%\\
 \textbf{CLACT} & 11.16 & 0.73 & 0.84 & 0.71 & 0.67  & -2.45\%\\
 \textbf{Amber-Pruner} & 10.64 & 0.74 & 0.84 & 0.70 & 0.64  & -1.23\%\\
 \textbf{LPTS} & 9.13 & 0.67 & 0.81 & 0.72 & 0.58  & 3.66\%\\
 \textbf{LPTS + VAR} & 9.10 & 0.68 & 0.81 & 0.73 & 0.56  & 3.97\%\\
 \textbf{R-SPARSE (64)} &9.03&0.79&0.76&0.75&0.64&-2.55\% \\
   \textbf{R-SPARSE (128)} &9.12&0.77&0.77&0.75&0.63&-1.51\% \\
%  \textbf{Amber-P qkv} & 97123 & 0.25 & 0.40 & 0.53 & 0.51 \\
% \textbf{Amber-P no\_qkv} & 10.64 & 0.74 & 0.84 & 0.70 & 0.64 \\

\midrule
%\multicolumn{6}{l}{\textbf{Gemma3-4B}} \\
\quad \textbf{Gemma3-4B-Instruct} & 17.29 & 0.72 & 0.84 & 0.72 & 0.62 & - \\ 
\midrule
 \textbf{ACT} & 35.62 & 0.65 & 0.76 & 0.70 & 0.51  & 9.94\%\\
 \textbf{WT} & 421.95 & 0.35 & 0.44 & 0.58 & 0.49  & 34.86\%\\
 \textbf{D-PTS} & 35.94 & 0.70 & 0.76 & 0.70 & 0.60  & 4.58\%\\
 \textbf{S-PTS} & 35.84 & 0.71 & 0.77 & 0.70 & 0.60  & 3.93\%\\
 \textbf{VAR} & 33.25 & 0.60 & 0.76 & 0.63 & 0.54  & 5.04\%\\
 \textbf{CLACT} & 39.22 & 0.66 & 0.74 & 0.67 & 0.59  & 8.01\%\\
 \textbf{Amber-Pruner} & 35.56 & 0.67 & 0.76 & 0.68 & 0.61  & 5.91\%\\
 \textbf{LPTS} & 19.55 & 0.65 & 0.73 & 0.70 & 0.55  & 9.19\%\\
 \textbf{LPTS + VAR} & 19.13 & 0.65 & 0.74 & 0.71 & 0.53  & 9.82\%\\

\textbf{R-SPARSE (64)} &17.04&0.69&0.76&0.69&0.60&5.17\% \\
\textbf{R-SPARSE (128)}&16.17&0.68&0.75&0.70&0.61&5.16\% \\
   
\bottomrule
\end{tabular}%
}
% \end{minipage}
\hfill

% \begin{minipage}[t]{0.49\textwidth}
\label{tab:main_two_four}
\centering
\end{table*}

\begin{table*}[hbt]
\centering
% \begin{minipage}[t]{0.49\textwidth}
\centering
\caption{\textbf{Semi-Structured 8:16 Sparsification} - performance Metrics, for calibration, when it is required, and perplexity we use WikiText2. Average Drop is computed without accounting for perplexity.}
\resizebox{!}{0.46\textheight}{%
\begin{tabular}{lcccccc}
\toprule
\textbf{Pruning} & \textbf{WikiText2} $\downarrow$ & \textbf{ARC Easy} & \textbf{BoolQ} & \textbf{PIQA} & \textbf{WinoGrande} & \textbf{Average Drop \%} \\
\midrule
\quad \textbf{Llama2-7B-chat}  & 6.94 & 0.74 & 0.80 & 0.76 & 0.66 &- \\
\midrule
   \textbf{ACT} & 8.12 & 0.69 & 0.75 & 0.73 & 0.63 & 5.37\%\\
   \textbf{WT} & 20.47 & 0.64 & 0.76 & 0.72 & 0.61 & 7.84\%\\
   \textbf{D-PTS} & 6.92 & 0.70 & 0.73 & 0.75 & 0.64 & 4.63\% \\
   \textbf{S-PTS} & 6.93 & 0.70 & 0.73 & 0.75 & 0.66 & 3.87\% \\
   \textbf{VAR} & 6.67 & 0.69 & 0.72 & 0.75 & 0.65 & 4.85\%\\
   \textbf{CLACT} & 6.54 & 0.71 & 0.74 & 0.75 & 0.64 & 3.98\% \\
   \textbf{CLACT + PTS} & 7.00 & 0.69 & 0.72 & 0.73 & 0.64 & 5.63\% \\
   \textbf{CLACT + VAR} & 6.72 & 0.69 & 0.73 & 0.75 & 0.64 & 5.07\% \\
   \textbf{R-SPARSE (64)} & 7.75 & 0.69 & 0.71 & 0.73 & 0.64 & 5.91\% \\
   \textbf{R-SPARSE (128)} & 7.82 & 0.68 & 0.69 & 0.74 & 0.61 & 7.93\% \\
   \textbf{Amber-Pruner} & 8.10 & 0.66 & 0.75 & 0.73 & 0.66 & 5.32\% \\
   \textbf{Amber-Pruner + PTS} & 6.90 & 0.68 & 0.72 & 0.72 & 0.65 & 6.16\% \\
   \textbf{Amber-Pruner + VAR} & 6.66 & 0.70 & 0.72 & 0.74 & 0.65 & 4.74\% \\
   \textbf{LPTS} & 7.50 & 0.69 & 0.66 & 0.74 & 0.63 & 8.15\% \\
   \textbf{LPTS + VAR} & 7.52 & 0.69 & 0.67 & 0.74 & 0.64 & 6.87\% \\

\midrule
\quad \textbf{Llama3.1-8B-Instruct} & 7.21  & 0.82 & 0.84 & 0.80 & 0.73 & - \\
\midrule
   \textbf{ACT} & 10.32 & 0.75 & 0.80 & 0.76 & 0.66 & 7.38\%\\
   \textbf{WT} & 22.56 & 0.51 & 0.64 & 0.63 & 0.54 & 27.26\%\\
   \textbf{D-PTS} & 10.34 & 0.76 & 0.80 & 0.76 & 0.66 & 6.79\%\\
   \textbf{S-PTS} & 10.31 & 0.76 & 0.80 & 0.75 & 0.66 & 7.30\%\\
   \textbf{VAR} & 10.67 & 0.74 & 0.79 & 0.75 & 0.66 & 8.30\%\\
   \textbf{CLACT} & 10.67 & 0.73 & 0.79 & 0.74 & 0.66 & 8.60\% \\
   \textbf{CLACT + PTS} & 10.68 & 0.74 & 0.79 & 0.74 & 0.65 & 8.55\% \\
   \textbf{CLACT + VAR} & 10.15 & 0.74 & 0.79 & 0.75 & 0.64 & 8.59\% \\
   \textbf{R-SPARSE (64)} & 11.42 & 0.75 & 0.77 & 0.75 & 0.66 & 8.44\% \\
   \textbf{R-SPARSE (128)} & 10.43 & 0.75 & 0.78 & 0.74 & 0.66 & 8.49\% \\
   \textbf{Amber-Pruner} & 10.16 & 0.73 & 0.80 & 0.75 & 0.68 & 7.13\% \\
   \textbf{Amber-Pruner + PTS} & 10.17 & 0.75 & 0.80 & 0.75 & 0.66 & 7.42\% \\
   \textbf{Amber-Pruner + VAR} & 9.94 & 0.74 & 0.80 & 0.75 & 0.64 & 8.39\% \\
   \textbf{LPTS} & 10.04 & 0.76 & 0.79 & 0.77 & 0.65 & 7.19\% \\
   \textbf{LPTS + VAR} & 10.26 & 0.77 & 0.78 & 0.76 & 0.66 & 7.15\% \\

\midrule
%\multicolumn{6}{l}{} \\
\quad \textbf{Qwen2.5-7B-Instruct} & 7.46 & 0.69 & 0.86 & 0.75 & 0.60 & - \\
\midrule
   \textbf{ACT}  & 8.61 & 0.66 & 0.87 & 0.73 & 0.53 & 4.38\% \\
   \textbf{WT} & 40.79 & 0.59 & 0.82 & 0.67 & 0.52 & 9.54\%  \\
   \textbf{D-PTS} & 8.61 & 0.80 & 0.87 & 0.77 & 0.68 & -8.28\%\\
   \textbf{S-PTS} & 8.84 & 0.80 & 0.86 & 0.76 & 0.67 & -7.24\%\\
   \textbf{VAR} & 11.91 & 0.69 & 0.72 & 0.75 & 0.65 & 1.93\%\\
   \textbf{CLACT} & 8.94 & 0.77 & 0.85 & 0.73 & 0.65 & -4.02\% \\
   \textbf{CLACT + PTS} & 8.94 & 0.77 & 0.86 & 0.83 & 0.67 & -5.06\% \\
   \textbf{CLACT + VAR} & 8.87 & 0.76 & 0.84 & 0.71 & 0.65 & -2.90\% \\
   \textbf{R-SPARSE (64)} & 8.12 & 0.82 & 0.79 & 0.77 & 0.69 & -6.90\% \\
   \textbf{R-SPARSE (128)} & 8.24 & 0.80 & 0.79 & 0.77 & 0.67 & -5.40\% \\
   \textbf{Amber-Pruner} & 8.80 & 0.77 & 0.86 & 0.74 & 0.69 & -6.20\% \\
   \textbf{Amber-Pruner + PTS} & 8.79 & 0.77 & 0.85 & 0.73 & 0.64 & -3.40\% \\
   \textbf{Amber-Pruner + VAR} & 8.73 & 0.75 & 0.85 & 0.74 & 0.65 & -3.60\% \\
   \textbf{LPTS} & 8.23 & 0.69 & 0.83 & 0.75 & 0.57 & 1.70\% \\
   \textbf{LPTS + VAR} & 8.21 & 0.70 & 0.83 & 0.73 & 0.56 & 2.80\% \\

\midrule
%\multicolumn{6}{l}{\textbf{Gemma3-4B}} \\
\quad \textbf{Gemma3-4B-Instruct} & 17.29 & 0.72 & 0.84 & 0.72 & 0.62 & - \\ 
\midrule
   \textbf{ACT} & 25.31 & 0.70 & 0.81 & 0.71 & 0.55  & 4.76\%\\
   \textbf{WT} & 198.53 & 0.39 & 0.60 & 0.62 & 0.52  & 26.11\%\\
   \textbf{D-PTS} & 25.17 & 0.70 & 0.81 & 0.71 & 0.54 & 5.16\%\\
   \textbf{S-PTS} & 25.40 & 0.75 & 0.82 & 0.74 & 0.63 & -1.54\%\\
   \textbf{VAR} & 23.93 & 0.75 & 0.81 & 0.73 & 0.65 & -1.87\%\\
   \textbf{CLACT} & 25.85 & 0.75 & 0.81 & 0.71 & 0.61 & 0.60\% \\
   \textbf{CLACT + PTS} & 26.03 & 0.73 & 0.81 & 0.70 & 0.63 & 0.50\% \\
   \textbf{CLACT + VAR} & 24.78 & 0.74 & 0.81 & 0.70 & 0.63 & 0.54\% \\
   \textbf{R-SPARSE (64)} & 15.39 & 0.76 & 0.80 & 0.74 & 0.64 & -1.36\% \\
   \textbf{R-SPARSE (128)} & 14.55 & 0.74 & 0.80 & 0.73 & 0.63 & -0.44\% \\
   \textbf{Amber-Pruner} & 25.11 & 0.74 & 0.82 & 0.70 & 0.63 & 0.08\% \\
   \textbf{Amber-Pruner + PTS} & 25.28 & 0.75 & 0.81 & 0.70 & 0.63 & 0.17\% \\
   \textbf{Amber-Pruner + VAR} & 23.97 & 0.74 & 0.81 & 0.71 & 0.64 & -0.16\% \\
   \textbf{LPTS} & 15.73 & 0.70 & 0.79 & 0.73 & 0.56 & 4.21\% \\
   \textbf{LPTS + VAR} & 15.68 & 0.71 & 0.79 & 0.72 & 0.57 & 3.41\% \\

\bottomrule
\end{tabular}}
% \end{minipage}
\hfill

% \begin{minipage}[t]{0.49\textwidth}
\centering
\label{tab:main_eight_sixteen}
\end{table*}

\begin{table*}[t]
\centering
\caption{Llama3.1-8B-Instruct with 8{:}16 activation sparsity. LS+L-PTS indicates Learnable Diagonal Scale + Learnable Shift, “Layers” indicates the subset of linear layers where the method was applied. Drop is computed without accounting for perplexity.}
\label{tab:ls_lpts_var_by_layers}
\resizebox{\textwidth}{!}{
\begin{tabular}{@{}llrrrrrrrrrrrrr@{}}
\toprule
\textbf{Method} & \textbf{Layers} & \textbf{PPL} & \textbf{BoolQ} & \textbf{WinoGrande} & \textbf{PIQA} & \textbf{ARC Easy} & \textbf{ARC Chal.} & \textbf{HellaSwag} & \textbf{OpenBookQA} & \textbf{RTE} & \textbf{MMLU} & \textbf{Lambada stand.} & \textbf{Lambada (OpenAI)} & \textbf{Drop \% $\downarrow$} \\
\midrule
\textbf{ORIGINAL} & -- & -- & 0.8391 & 0.7340 & 0.8003 & 0.8207 & 0.5196 & 0.5905 & 0.3420 & 0.6859 & 0.6790 & 0.6569 & 0.7308 & -- \\
\midrule
LS+L-PTS & all            & 9.6036 & 0.7841 & 0.6638 & 0.7715 & 0.7647 & 0.4514 & 0.5294 & 0.2720 & 0.6318 & 0.5521 & 0.5686 & 0.6625 & 10.90\% \\
LS+L-PTS & k,o,gate,down  & 8.3483 & 0.8205 & 0.7111 & 0.7889 & 0.7887 & 0.4659 & 0.5591 & 0.3200 & 0.6462 & 0.6060 & 0.6123 & 0.7046 & 5.43\% \\
LS+L-PTS & k,v,gate,down  & 8.0821 & 0.8352 & 0.7174 & 0.7867 & 0.7992 & 0.4898 & 0.5651 & 0.3260 & 0.6643 & 0.6322 & 0.6262 & 0.7108 & 3.56\% \\
\midrule
LS+L-PTS + VAR & all           & 9.4983 & 0.7872 & 0.6606 & 0.7606 & 0.7601 & 0.4334 & 0.5372 & 0.2880 & 0.6390 & 0.5532 & 0.5729 & 0.6689 & 10.60\% \\
LS+L-PTS + VAR & k,o,gate,down & 8.2930 & 0.8116 & 0.7135 & 0.7851 & 0.7908 & 0.4838 & 0.5634 & 0.3300 & 0.6498 & 0.6095 & 0.6189 & 0.7079 & 4.64\% \\
LS+L-PTS + VAR & k,v,gate,down & 8.0259 & 0.8306 & 0.7269 & 0.7851 & 0.7955 & 0.4863 & 0.5673 & 0.3260 & 0.6715 & 0.6327 & 0.6317 & 0.7143 & 3.36\% \\
\bottomrule
\end{tabular}}
\end{table*}

% \begin{figure*}
%     \centering
%     \includegraphics[width=1\linewidth]{Figures/ACT_vs_WT_v4.png}
%     \caption{Comparison of \textbf{unstructured} \textbf{sparsity} in \textbf{activations} (blue) and \textbf{weights} (orange) across datasets at varying sparsity ratios. Dotted lines (BASE) denote the non-sparse model's performance. Accuracy is reported for all tasks except WikiText, where we use perplexity.}
%     \label{fig:enter-label}
% \end{figure*}

\section{Comparison with Quantization Baselines}
\label{app:quantization}
 
A key question for practitioners is whether activation sparsity offers competitive accuracy retention compared to quantization, the dominant compression technique in production LLM serving.
In Table~\ref{tab:quant_comparison}, we compare our post-training activation sparsity results against quantization baselines from \citet{zhelnin2025gift} on Llama3.1-8B-Instruct.
 
\begin{table*}[h]
\centering
\caption{Comparison of activation sparsity and quantization on Llama3.1-8B-Instruct. Quantization results are from \citet{zhelnin2025gift}. Note that GIFT-SW uses stochastic fine-tuning (STE), whereas our activation sparsity methods require \textbf{no fine-tuning}.}
\label{tab:quant_comparison}
\small
\begin{tabular}{lcccc}
\toprule
\textbf{Method} & \textbf{BoolQ} & \textbf{WinoGrande} & \textbf{PIQA} & \textbf{ARC-Easy} \\
\midrule
Baseline (dense)             & 0.839 & 0.734 & 0.800 & 0.821 \\
\midrule
\multicolumn{5}{l}{\textit{Quantization (requires fine-tuning)}} \\
8-bit GIFT-SW (STE)          & ---   & 0.738 & 0.810 & 0.798 \\
\midrule
\multicolumn{5}{l}{\textit{Activation sparsity (no fine-tuning)}} \\
50\% unstruct.\ + S-PTS       & 0.800 & 0.660 & 0.750 & 0.760 \\
50\% unstruct.\ + VAR         & 0.819 & 0.705 & 0.776 & 0.784 \\
8:16 + ACT (magnitude)       & 0.797 & 0.655 & 0.757 & 0.753 \\
8:16 + Amber-Pruner          & 0.800 & 0.680 & 0.750 & 0.730 \\
8:16 + D-PTS                 & 0.800 & 0.660 & 0.760 & 0.760 \\
8:16 + VAR                   & 0.790 & 0.660 & 0.750 & 0.740 \\
\bottomrule
\end{tabular}
\end{table*}
 
Several observations emerge from this comparison.
First, 8-bit quantization with fine-tuning (GIFT-SW~\cite{zhelnin2025gift}) achieves strong results, in some cases exceeding the dense baseline (e.g., WinoGrande: $0.738$ vs.\ $0.734$), but it requires gradient-based stochastic training, which incurs significant computational cost and risks degrading safety alignment~\citep{kharinaev2025investigating}.
In contrast, our activation sparsity methods are entirely \textbf{post-training} and require \textbf{no fine-tuning}, making them immediately deployable without retraining infrastructure.
 
Second, unstructured 50\% activation sparsity with VAR achieves competitive performance across all four benchmarks, with an average drop of only $3.47\%$, while providing a $2\times$ theoretical FLOP reduction.
The semi-structured 8:16 variants incur somewhat larger drops ($6$--$8\%$), but offer hardware-friendly regularity that can be exploited by future accelerators. Finally, we note that this comparison is limited to a single model and a subset of benchmarks.
A comprehensive comparison would require evaluating additional quantization methods (GPTQ~\cite{frantar2022gptq}, AWQ~\cite{lin2024awq}, SqueezeLLM~\cite{kim2023squeezellm}). 
\end{document}